\documentclass[letterpaper, 10 pt, conference]{ieeeconf}  % Comment this line out
\IEEEoverridecommandlockouts % This command is only
\usepackage{graphicx} % for pdf, bitmapped graphics files
\graphicspath{ {Figures/} }
\usepackage{amsmath} % assumes amsmath package installed
\usepackage{amssymb}  % assumes amsmath package installed
\usepackage{bm}
\newcommand{\inv}{^{\raisebox{.2ex}{$\scriptscriptstyle-1$}}}
\usepackage{gensymb}

\usepackage{color,soul} % highlight
\usepackage{booktabs}
\usepackage{hyperref} % url

\title{\LARGE \bf
Robotic Sewing and Knot Tying for Personalized Stent Graft Manufacturing 
}

\author{Yang Hu$^{*}$, Lin Zhang$^{*}$, Wei Li and Guang-Zhong Yang,~\IEEEmembership{Fellow,~IEEE}% <-this % stops a space
\thanks{$^{*}$ These authors contributed equally to the work.}%
\thanks{All authors are with The Hamlyn Centre for Robotic Surgery, Institute of Global Health Innovation,	Imperial College London, SW7 2AZ, UK}%
}

\begin{document}

\maketitle
\thispagestyle{empty}
\pagestyle{empty}

%%%%%%%%%%%%%%%%%%%%%%%%%%%%%%%%%%%%%%%%%%%%%%%%%%%%%%%%%%%%%%%%%%%%%%%%%%%%%%%%
\begin{abstract}
This paper presents a versatile robotic system for sewing 3D structured object. Leveraging on using a customized robotic sewing device and closed-loop visual servoing control, an all-in-one solution for sewing personalized stent graft is demonstrated. Stitch size planning and automatic knot tying are proposed as the two key functions of the system. By using effective stitch size planning, sub-millimetre sewing accuracy is achieved for stitch sizes ranging from 2mm to 5mm. In addition, a thread manipulator for thread management and tension control is also proposed to perform successive knot tying to secure each stitch. Detailed laboratory experiments have been performed to access the proposed instruments and allied algorithms. The proposed framework can be generalised to a wide range of applications including 3D industrial sewing, as well as transferred to other clinical areas such as surgical suturing.  

\end{abstract}

%%%%%%%%%%%%%%%%%%%%%%%%%%%%%%%%%%%%%%%%%%%%%%%%%%%%%%%%%%%%%%%%%%%%%%%%%%%%%%%%
\section{Introduction}

Vascular disease is a major contributor to cardiovascular deaths in the world. The incidence of abdominal aortic aneurysms increases significantly with age-by over 300\% for those at age 70 compared to those at age 50. Endovascular therapy avoids major trauma associated with open operation, with clear advantages in terms of reduced morbidity and mortality, especially for patients unable to withstand traditional open surgery. This procedure of endovascular intervention requires the use of personalized stent grafts with bespoke openings (fenestrations) and branches for small blood vessels, thus providing better fit for a patient's anatomy. These personalized stent grafts are currently handmade by means of sewing thousands of stitches, which are expensive and time-consuming. The long delay in customized graft manufacturing can subject patients to the risk of aneurysm rupture and precludes treatment for patients with acute symptoms. The development of an automated stent graft sewing technique would be very helpful to increase the speed of personalized stent manufacturing.

\begin{figure}[t]
	\centering
	\includegraphics[width=\linewidth]{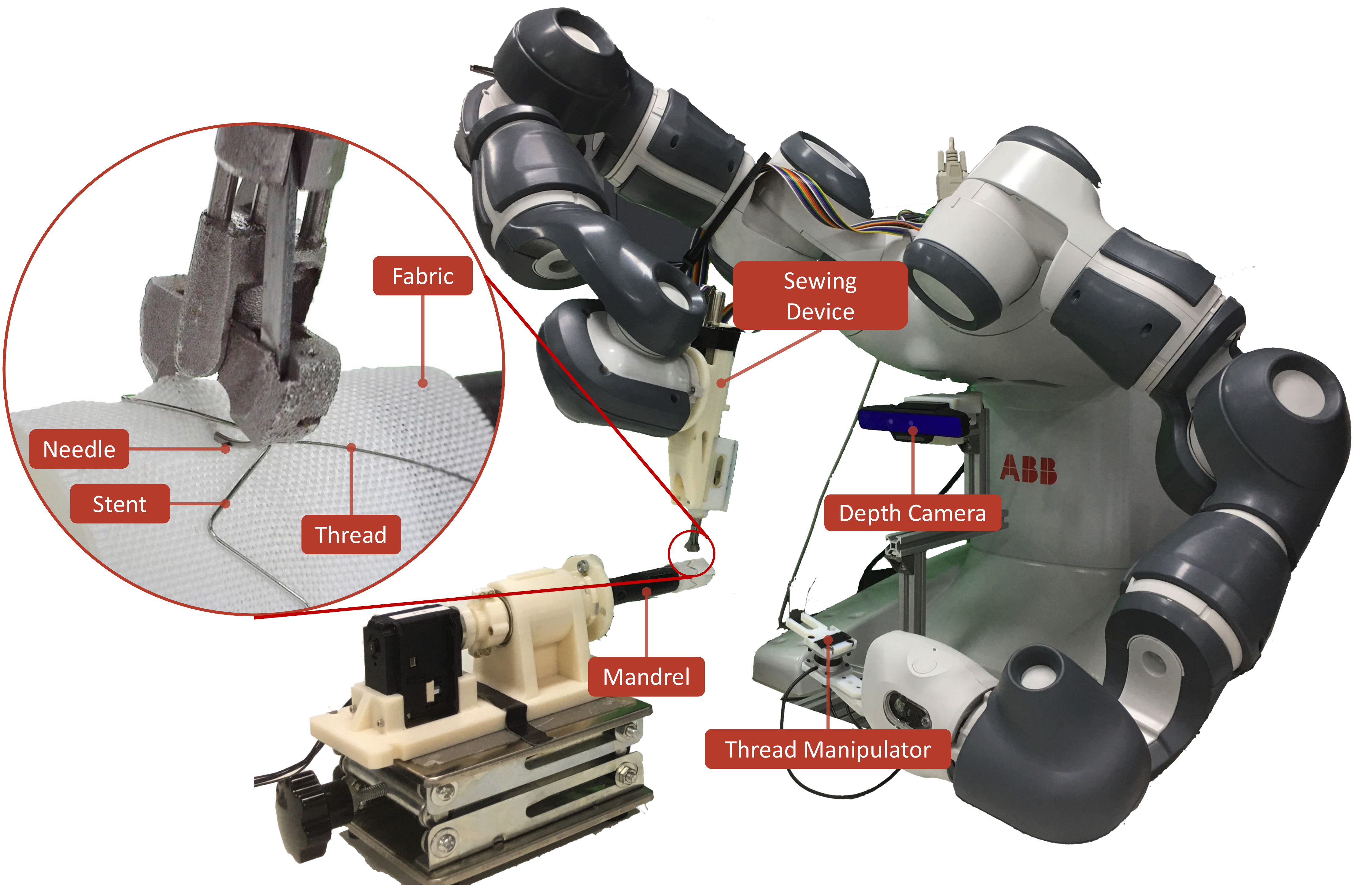}
	\caption{An overview of the setup of the autonomous sewing system for stent graft manufacturing.}
	\label{fig:intro}
\end{figure}

Thus far, a few automated sewing methods have been proposed, e.g., a modified sewing machine is used for sewing tubular shape stent grafts as proposed by \cite{phillips2006stent}. A bobbin, which is seated in a shuttle, carries a bottom thread through the bore of the tubular graft and forms a stitch in combination with a top thread carried on a needle which pierces the graft wall. This solution, however, has a number of potential issues. First, these sewing system is usually specific for one type of stitch; therefore it is not versatile for sewing a custom-made stent graft, which requires running stitch to attach the stent with the graft, applying blanket stitch to finish the edge of a fenestration and even tying knots the secure the sewing. Second, compared with handmade stitch, which only requires using one thread, double thread machine stitches are thick and easily coming out from the fabric once any point breaks. In order to solve the above problem, a vision-guided robotic sewing system for stent graft manufacturing is proposed by Huang et al. \cite{huang2016stent, huang2017multi}. This system mimics human sewing method by learning a specific sewing method from human demonstrations. One limitation of this system is that the success rate is only 75\%. We believe the low success rate is mainly because this system uses two robot arms, each holding a grasper to manipulate a curved needle. Because the uncertainty in needle manipulation, e.g. during piercing and re-grasping, the needle might drop and slip from the grasper. In addition, this method uses a look and move method to guide the sewing process, which is slow, and the motion is not smooth, making it difficult to perform delicate stitching with high accuracy. In order to solve this problem, a multi-purpose sewing device was developed by us and patented for a wide range of applications including 3D structured object sewing, as well as transferred to other clinical areas such as surgical suturing.
This device could switch a double pointed curved needle to perform sewing. The thread is attached to the middle of the needle. The device avoids the problem of needle slippage and dropping in manual sewing. The nominal speed to complete one needle switching is 16s, which shows faster, reliable and more consistent performance than directly operating a surgical needle with a needle holder, making this device a suitable candidate for fully automated robotic sewing. The reader can refer to our following publications for the mechanical design of this new device. In this paper, we propose a robotic sewing system (Fig.~\ref{fig:intro}) integrated with the sewing device and depth sensing, which could automate some sewing tasks, like running stitch and knot tying. The contribution of this work is as follows:
\begin{itemize}
  \item A dual-arm robotic system for sewing stent grafts. The robot arm is used for global positioning the sewing device, and the sewing device is used for local needle manipulation. 
  \item Application of a closed-loop visual servoing method for accurate positioning of the sewing device.
  \item A trajectory planning method is designed to perform stitches in various sizes.
  \item A novel knot tying method including a thread manipulator is proposed, which can continuously make overhand knots without cutting the thread.
\end{itemize}

The remainder of the paper is organized as following. We first present the prior art and related work in Section \ref{sec:relate_work}. Section \ref{sec:methods} introduces the three new features of our system, i.e. the vision-guided stitching, stitch size planning and automatic knot tying. Section \ref{sec:exp_result} shows the experiment results, followed by the results discussion in section \ref{sec:discuss}. Then finally comes the conclusion. 

\section{Related Works}
\label{sec:relate_work}
Extensive research on automated sewing has been performed in the apparel industry focusing on incorporating sensors and robots to augment the ability of conventional sewing machines. Relevant topics include fabric tension control for robot-assisted fabric feeding \cite{schrimpf2012sew}, sewing seam tracking using an optical sensor \cite{schrimpf2012multi} or a camera system \cite{kudo2000multi}, multi-arm robotic sewing \cite{schrimpf2012experiments}, and automatic sewing worklines \cite{guizzo2018sewbot}. These systems, however, constrained by the incorporation of conventional sewing machines, work only on a flat sewing table. Recently, innovation in 3D structured object sewing is an important topic for industrial manufacturing. 
KSL Keilmann (Lorsch, Germany) has developed different single-sided sewing heads, such as KSL double needle RS 530, for sewing 3D fabric-reinforced structures for aircraft parts. Different to a conventional sewing machine, which applies stitches by using synchronized parts working on both side of the fabric, the single-sided sewing head features two needles, which are able to apply interwoven chain stitches from the outer surface of a 3D object. These single-sided sewing heads, however, are designed to sew large and heavy objects. 

The use of thread to stitch together objects is not only employed in the field of the apparel industry. Automated suturing in the field of robotic surgery is also a hot topic which arouses researchers' attention. Form the software side, a number of works, such as \cite{jackson2013needle,sen2016automating} have been trying to plan the needle trajectory to achieve optimal stitching result. On the hardware side, various suturing devices aiming to facilitate the suturing procedure have been developed, including the Autosuture EndoStitch\textsuperscript{TM} (Covidien, Mansfield, UK) \cite{gopel2011automation}, the SILS\textsuperscript{TM} Suturing Device (Covidien, Mansfield, UK) \cite{adams1995suture}, Endo360\textsuperscript{o} Suturing Device, (EndoEvolution North Chelmsford, USA) \cite{ meade2011apparatus}, the PROXISURE\textsuperscript{TM} Suturing Device (Ethicon, New Jersey, USA) \cite{martin2016circular} and the Switch\textsuperscript{\textregistered} Suturing Device (Mellon Medical, Netherlands) \cite{holwerda2014surgical}. 
% According to the ways to operate the needle, these devices can be classified into two categories. For the first category, the suturing devices, e.g. Switch\textsuperscript{\textregistered}, EndoStitch\textsuperscript{TM} and SILS\textsuperscript{TM} systems switch a double pointed needle between two opposing jaws. A locking mechanism is built in each jaw to lock the needle. For the second category, the suturing devices, e.g. Endo360\textsuperscript{o} and PROXISURE\textsuperscript{TM}, work by continuously rotating a circular needle to perform stitching. 
Based on the aforementioned suturing devices, vision-guided robotic suturing systems, such as KidsArm Anastomosis Robot proposed by Looi et al. \cite{looi2013kidsarm} and Smart Tissue Autonomous Robot (STAR) proposed by Leonard et al. \cite{leonard2014star, shademan2016sci}, demonstrate that it is feasible for a robot to autonomously perform anastomosis. Knot tying, used for securing the stitches or binding together objects is another important part of the robotic surgery. Making a knot in the confined space of  minimally invasive surgery under poor visualization is a very challenging surgical task. The methods of automated knot tying have been proposed for reducing the burden of the surgeon. With using the conventional surgical instruments, a general way to tie a knot is conducted by using one grasper to wind the suture thread around another grasper to form a loop and then used the first grasper to pull the tail end of the suture through the loop. To achieve automated knot tying, various methods were proposed, e.g. thread shape modelling \cite{yue2007twisting}, instrument moment planning \cite{chow2013improved}, and learning from surgeons' demonstration \cite{mayer2008system}. Another direction for automated knot tying is by designing specific hardware, e.g. \cite{jernigan2010laparoscopic} proposed a knot tying device for minimally invasive surgery, \cite{bell2008knot} proposed a single piece fixture for knot tying, and \cite{kang2002robotic} uses the Endostitch\textsuperscript{TM} suturing device and a grasper with a flexible hook to make a knot. 
 
\section{Methodology}
\label{sec:methods}

\subsection{System Overview}
\begin{figure}[t]
	\centering
	\includegraphics[width=\linewidth]{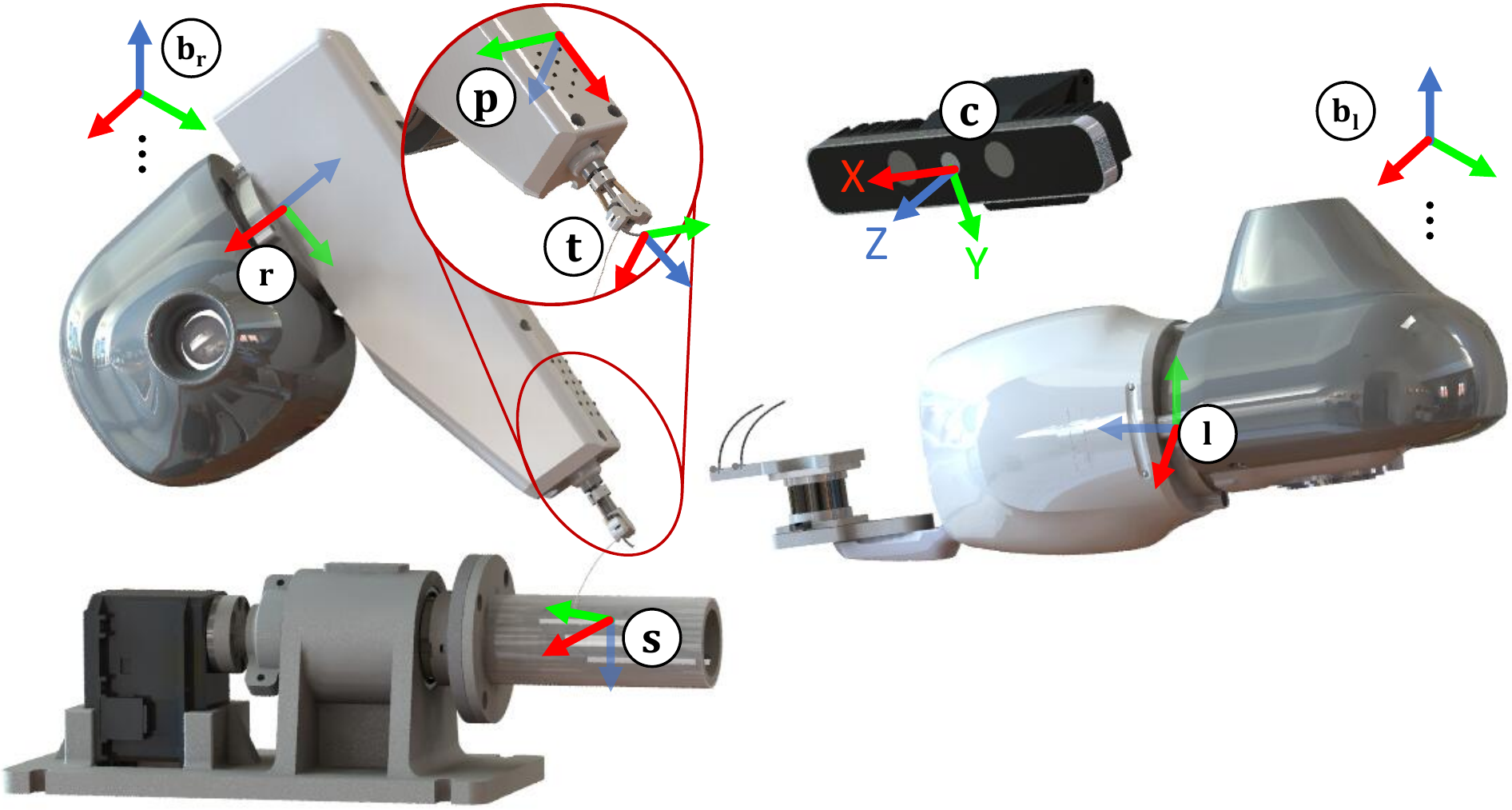}
	\caption{Illustration of the coordinate frames used in this paper. The camera coordinate frame is located at the optical centre of the colour camera which is denoted as \{$\mathbf{c}$\}. The pattern and needle tip coordinate frames are denoted as \{$\mathbf{p}$\} and \{$\mathbf{t}$\} respectively. The left and right robot base coordinate frames are denoted as \{$\mathbf{b}_l$\} and \{$\mathbf{b}_r$\} respectively. The coordinate frame at the sewing point on the fabric is denoted as \{$\mathbf{s}$\}. The robot left and right end-effector coordinate frames are denoted as \{$\mathbf{l}$\} and \{$\mathbf{r}$\} respectively.}
	\label{fig:coord_def}
\end{figure}

\begin{figure}[tb]
	\centering
	\includegraphics[width=\linewidth]{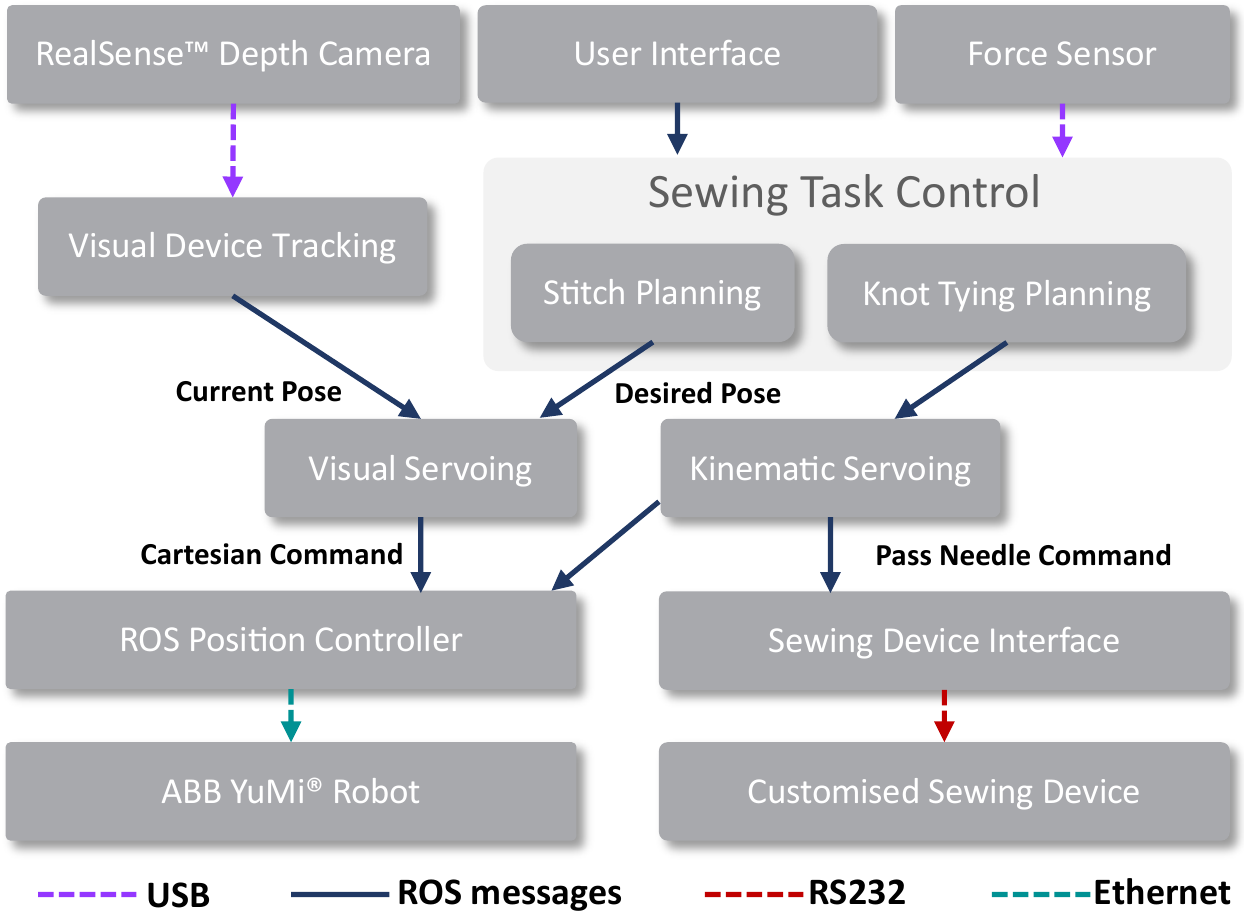}
	\caption{An overview of the framework developed in this paper.}
	\label{fig:framework}
\end{figure}

The system consists of a dual-arm ABB YuMi\textsuperscript{\textregistered} robot where each arm has 7 DoF. The end-effector of the left and right arm is equipped with a thread manipulator and a customised sewing device, respectively. The dual-arm robot is controlled using the ROS controller manager and the ABB Robot Web Services (RWS) via the Ethernet connection. The RWS interface provides a bandwidth of 10 Hz between the host computer and the robot. A joint position controller\footnote{https://github.com/kth-ros-pkg/yumi} is used for the robot position control.

An Intel\textsuperscript{\textregistered} RealSense\textsuperscript{TM} SR300 depth camera is rigidly mounted under the robot as shown in Fig.~\ref{fig:intro}, which is used to capture colour and depth images of the scene for automating the sewing task. The resolution of both colour and depth images are 640x480 and the intrinsic and extrinsic parameters of the colour and depth camera are provided by the manufacturer. The point cloud generated by the depth camera is in the camera coordinate. The robot base coordinate is registered with the camera reference frame via standard hand-eye calibration \cite{tsai1989handeye}. A complete definition of the coordinate frame used in this paper is shown in Fig.~\ref{fig:coord_def}.

In this paper, a patented sewing device developed by the authors is used. The device features a sewing probe which can pass a double pointed semi-circular needle between two movable jaws to perform sewing.  It has a needle driving mechanism that can drive the needle along its tangential direction. It consists of a needle locking mechanism built in each jaw to allow needle holding. The device simplifies the needle manipulation by providing fast and repeatable needle grasping and releasing. A marker with dot pattern is attached to the sewing device to facilitate tracking of the sewing device. 

\subsection{Visual Servoing for Accurate Positioning}
In the proposed system, Image-Based Visual Servoing (IBVS) is applied for accurate positioning of the sewing device. For each point on the pattern, its 3-D coordinate in the camera coordinate frame \{$\mathbf{c}$\} can be written as $\bm{X_c}=[X, Y, Z,1]^\intercal$, which has a corresponding 2D projection $\mathbf{m}=[u, v]^\intercal$ on the image plane in pixel units.
The $\mathbf{m}$ can be converted to normalised coordinate $\mathbf{x} = [x, y]^\intercal$ in metre via:
\begin{equation}
\label{eq:norm_xy}
  \begin{cases}
  	x=\frac{u-c_u}{fx} \\
  	y=\frac{v-c_v}{fy} \\
  \end{cases}
\end{equation}
where $f_x$, $f_y$ is the camera focal length expressed in the pixel units and $c_u$, $c_v$ is the camera principal point.

 The velocity of the 3-D point relative to \{$\mathbf{c}$\} is $\mathbf{v_c}=[v_x, v_y, v_z, \omega_x, \omega_y, \omega_z]^\intercal$, with $v$ the instantaneous linear velocity and $\omega$ the instantaneous angular velocity. The velocity of the corresponding $i$th 2D point $\dot{\mathbf{x}}$ can be derived via:
\begin{equation}
\label{eq:x_dot}
  \dot{\mathbf{x}}=\mathbf{L_{x_i}}\mathbf{v_c}
\end{equation}
where $\mathbf{L_{x_i}}$ an interaction matrix that is written as:
\begin{equation}
\label{eq:L_x}
  \mathbf{L_{x_i}} = \begin{bmatrix}
-\frac{1}{Z} & 0 & \frac{x}{Z} & xy & -(1+x^2) & y \\
0 & -\frac{1}{Z} & \frac{y}{Z} & 1+y^2 & -xy & -x
\end{bmatrix}
\end{equation}

The depth $Z$ is obtained from the RGBD camera. 
The 2D point on the image is tracked by a real-time vision-based method \cite{zhang2017track}.

During the task execution, the pattern is tracked in each image frame while the dots in the pattern with invalid depth value are discarded. The $n$ valid points are used to construct the interaction matrix by stacking (\ref{eq:L_x}):
$$
  \mathbf{L_{x}} = \begin{bmatrix}
 	\mathbf{L_{x_1}} \\
 	\mathbf{L_{x_2}} \\
 	\vdots					\\
 	\mathbf{L_{x_n}}
 \end{bmatrix}.
$$

The 2D velocity $\dot{\mathbf{x}}$ of a point can be written as 
\begin{equation}
\label{eq:x_dot}
  \dot{\mathbf{x}}=\mathbf{x}-\mathbf{x^*}
\end{equation}
where $\mathbf{x^*}$ is the desired position of the point.
To transform the velocity $\mathbf{v_c}$ from the camera coordinate to the robot end-effector coordinate $\mathbf{v_e}$, a velocity twist $^e\mathbf{V}_c$ is applied: $\mathbf{v_e}=\mathbf{^{e}V_c}\mathbf{v_c}$. A velocity twist $\mathbf{^{k}V_j}$ that transforms velocities from coordinate $\{\mathbf{j}\}$ to coordinate $\{\mathbf{k}\}$ is defined as:

$$
\mathbf{^{k}V_j} = \begin{bmatrix}
\mathbf{^{k}R_j} & [\mathbf{^{k}t_j}]_{\times} \mathbf{^{k}R_j} \\
\mathbf{0_{3\times3}} & \mathbf{^{k}R_j}
\end{bmatrix}.
$$

The $\mathbf{^{k}R_j}$ and $\mathbf{^{k}t_j}$ is the rotation matrix and translation vector that transform points from $\{\mathbf{j}\}$ to $\{\mathbf{k}\}$. The $[\mathbf{^{k}t_j}]_{\times}$ is the skew-symmetric matrix of the translation vector. In the proposed system, the transformation ($\mathbf{^{b}R_c}$ and $\mathbf{^{b}t_c}$) between the camera coordinate $\{\mathbf{c}\}$ and robot base coordinate $\{\mathbf{b}\}$ is obtained from the hand-eye calibration. The transformation between the robot base $\{\mathbf{b}\}$ and end-effector $\{\mathbf{e}\}$ is computed by using the robot forward kinematics. The velocity twist $\mathbf{^{e}V_c}$ can be written as:

\begin{equation}
\label{eq:eVc}
\mathbf{^{e}V_c} = \mathbf{^{e}V_b} \mathbf{^{b}V_c}	
\end{equation}

Using (\ref{eq:x_dot}) and (\ref{eq:eVc}), the desired end-effector velocities can be calculated by:
\begin{equation}
\label{eq:v_e}
\mathbf{v_e} = (\mathbf{L_x} \mathbf{^{c}V_b} \mathbf{^{b}V_e})\inv \dot{\mathbf{x}}
\end{equation}

\subsection{Trajectory Planning for Automated Stitching}

\begin{figure*}[t]
	\centering
	\includegraphics[width=\linewidth]{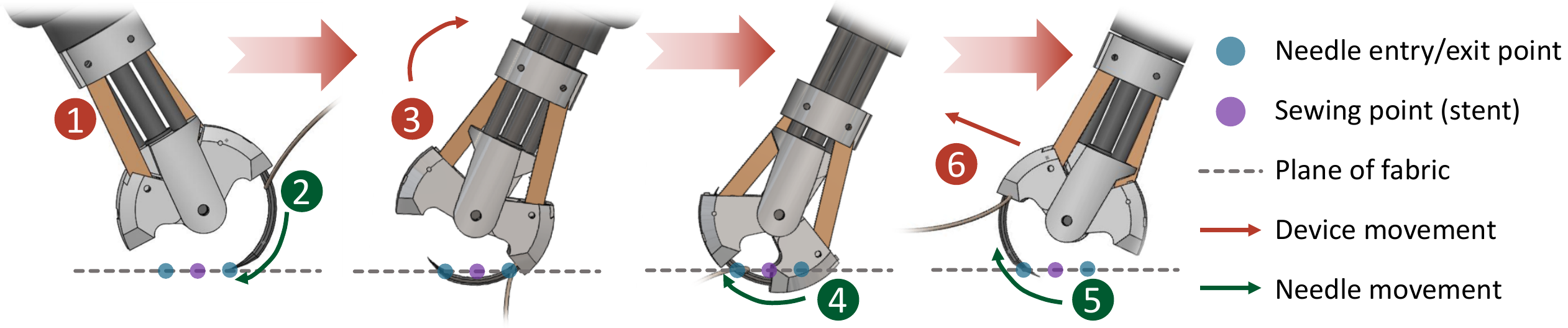}
	\caption{Illustration of a stitching process using the customised sewing device. For a single stitch, six steps (indicated by the number) are needed to accomplish the stitch at a target (purple point) with a desired stitch size (distance between two blue points).}
	\label{fig:stitch_steps}
\end{figure*}

The planned sewing trajectory consists of a sequence of needle tip poses $\mathbf{^{c}T_t}$ in \{$\mathbf{c}$\}. As the visual servoing method uses the projection of the pattern dots for robot control, the desired 3D dot positions in $\{\mathbf{c}\}$ are required:
\begin{equation}
\label{eq:X_c^*}
  \bm{X_c}^* = \mathbf{^{c}T_t} \mathbf{^{t}T_p} \bm{X_p},
\end{equation}
where $\mathbf{^{t}T_p}$ is a fixed transformation between the needle tip and pattern that is given by the CAD model of the device. The desire robot end-effector velocity $\mathbf{v_e}^*$ can then be calculated using equation (\ref{eq:norm_xy}, \ref{eq:x_dot}, \ref{eq:v_e} and \ref{eq:X_c^*}).

The stitching task can be accomplished via six steps as illustrated in Fig.~\ref{fig:stitch_steps}. The fabric piercing (step 2 in Fig.~\ref{fig:stitch_steps}), needle switching (step 4 in Fig.~\ref{fig:stitch_steps}) and retrieving (step 5 in Fig.~\ref{fig:stitch_steps}) are handled by the sewing device. To complete one stitch with desired size, the robot arm performs the other three steps with three planned trajectories with respect to the stitching point: (a) a trajectory for moving the device from the current position to a sewing target for stitching (step 1 in Fig.~\ref{fig:stitch_steps}); (b) a trajectory for reorienting the device so that the needle can be passed to the other side easier (step 3 in Fig.~\ref{fig:stitch_steps}); (c) a trajectory for pulling the needle out of the fabric (step 6 in Fig.~\ref{fig:stitch_steps}).

The first part of the stitching trajectory is planned according to the 3D profile of the fabric. A 3D point cloud is obtained from the RGBD camera. The normal to a selected target point on the fabric is estimated by computing the local gradient of the neighbouring points. As shown in Fig.~\ref{fig:coord_def}, the objective is to align the local needle tip frame \{$\mathbf{t}$\} with the local sewing frame \{$\mathbf{s}$\}. More specifically, the desired needle tip pose $\mathbf{^{c}T_t}$ is planned so that the local z-axis is pointing to the surface along the normal. The direction of x-axis is defined by the direction of the current and the next sewing points. The y-axis is the cross-product of the x- and y-axis. To allow easy needle piercing, the device is oriented $-\alpha$ around the tip local x-axis so that the needle tip is more perpendicular to the fabric surface. In addition, the stitch point has an offset from the sewing point which should be the half of the stitch size $d$. The above information can be applied to the desired tip pose via 
\begin{equation}
\label{eq:cTt*}
  \mathbf{^{c}T_{t^*}}=\mathbf{^{c}T_t \mathbf{^{t}T_{t^*}}}
\end{equation}
where $\mathbf{^{t}T_{t^*}}$ is defined as:
$$
\mathbf{^{t}T_{t^*}}=\begin{bmatrix}
 	1 & 0 & 0 & 0 \\
 	0 & \cos\alpha & -\sin\alpha & \frac{d}{2}	\\
	0 & \sin\alpha & \cos\alpha & 0		\\
	0 & 0 & 0 & 1
 \end{bmatrix}.
$$

After the needle arrives at the entry point, as depicted by step 2 in Fig.~\ref{fig:stitch_steps}, the needle pierces into the fabric for $l$ which is defined as $l=\theta R$. $\theta$ is the angle that the jaws are rotated and $R$ is the radius of the needle. The length of the pierced segment of the needle equals to the stitch size that is $d=l$.

Next, the device is reoriented so that the needle can make contact with the exit point on the fabric, as depicted by step 3 in Fig.~\ref{fig:stitch_steps}. A rotation of $2\alpha$ is applied to the device similar to (\ref{eq:cTt*}). To minimise the force applied on the fabric by the needle, the needle should rotate around the entry point without stretching the fabric. To this end, a sequence of intermediate poses is planned by rotating the needle 5\degree each time. Upon arrival, the needle is passed to the other jaw to accomplish the stitch which is shown in step 5 of Fig.~\ref{fig:stitch_steps}.

After a stitch is done, a trajectory is planned to pull the needle and thread out of the fabric, as depicted by step 6 in Fig.~\ref{fig:stitch_steps}. Interpolate between the current needle tip pose and a desired needle pose at a standby position is performed. More specifically, linear interpolation and spherical linear interpolation are used for translation and rotation, respectively.

\subsection{Automatic knot tying}

\begin{figure*}[t]
	\centering
	\includegraphics[width=\linewidth]{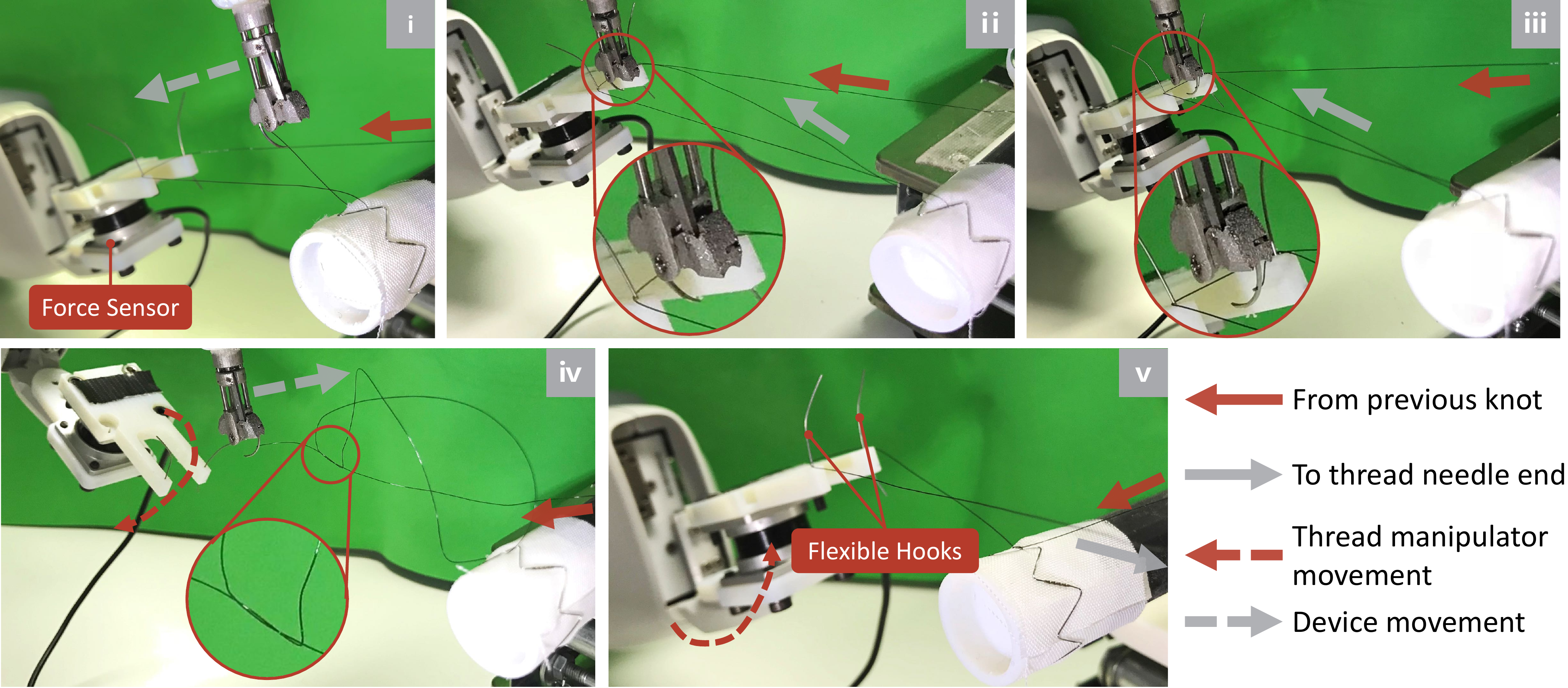}
	\caption{An illustration of the proposed knot tying method using the thread manipulator.}
	\label{fig:knot_steps}
\end{figure*}

In this paper, a knot tying method is proposed to tie an overhand knot. This is accomplished by a combined used of the sewing device and a novel thread manipulation mechanism, called thread manipulator. Different with previous methods, one unique feature of the device is that it can perform successive stitching and knot tying along the sewing trajectory without cutting or replacing the thread. Making a loop and passing the thread end through the loop are two critical procedures for knot tying. The thread manipulator is introduced for three purposes: assisting in making a loop, keeping the thread in a known position and preventing the thread from floating everywhere, which may hinder the sewing process. The two flexible hooks attached to the thread manipulator can assist to make a loop. By moving the tip of the sewing device around the two flexible hooks, followed by the needle end of the thread, a loop is made. The reason for using two flexible hooks is that if the thread segment between the two flexible hooks is pulled tightly, the sewing device could easily pass the needle around the fixed thread segment to make an overhand knot.

To control the robot to carry out the knot tying task, a sequence of key poses are recorded via human demonstration. These poses are in the relative frame between the robot left end-effector frame and right end-effector frame, which is denoted as $\mathbf{^{l}T_r}$. During the knot tying task, current left end-effector pose $\mathbf{^{\mathbf{b}_l}T_l}$ is obtained from the robot kinematics, then the desired right end-effector pose can be computed via $\mathbf{^{\mathbf{b}_r}T_r} = \mathbf{^{\mathbf{b}_l}T_l}\mathbf{^{l}T_r}$. It should be noted that the base coordinate frame of the left and right arms are identical (i.e. $\mathbf{^{\mathbf{b}_l}T_{\mathbf{b}_r}}=\mathbf{I}_4$).

In order to control the tension of the thread, the two flexible hooks are connected to a 3-axis force sensor (OptoForce Ltd, Budapest, Hungary), which is mounted on the end-effector of the left robot arm. In the tension control mode, instead using a fixed relative pose between the two arm, the sewing device arm serves as the leader to pull the thread while the thread manipulator arm serves as the follower to feed in the thread. By using the force measurement along the dominant axis as feedback and controlling the position of the thread manipulator arm, the tension of the thread can be kept with a simple proportional control strategy.
 
	%displacement equal gain(desired_force-measured_force)
    
Our knot tying method is a recursive method which contains five steps, as illustrated in Fig.~\ref{fig:knot_steps}. We assume a knot is already formed using this method and the following describes how to make next knot. The initial condition is that the sewing device taking the thread move around the flexible hook, using the force sensor's reading to secure the previous knot. Start from the accomplishment of the stitching task, it takes following five steps to finish a knot:

\begin{itemize}
\item As a stitch has been made by the right arm using the sewing device, the left hand start pulling the thread to ensure tension until a force threshold is achieved. In the meantime, the right arm moves towards the standby position relative to the left hand.
\item The sewing device moves to catch the thread segment between the two flexible hooks. At the same time, the tension control is turned on. By sensing the force on the flexible hooks, the left arm starts feeding in the thread with a predefined tension.
\item After reaching the position where it can catch the fixed thread segment within its jaws, the sewing device perform a needle switching. Then, an overhand knot is made.  
\item The assistant arm tilts the flexible hooks to release the knot. 
\item The sewing device takes the thread move around the flexible hook, using the force feedback to secure the previous knot. 
\end{itemize}

\section{Experiments and Results}
\label{sec:exp_result}

\begin{table}[!b]
\centering
\caption{The results of running stitch with different sizes.}
\label{tb:stitch}
\begin{tabular}{@{}ccccccc@{}}
\toprule
 &1mm&2mm&3mm&4mm&5mm\\ \midrule
Trial 1         & 2.0     & 2.5     & 3.5     & 4.5     & 5.5 \\
Trial 2         & 2.5     & 2.0     & 3.3     & 3.5     & 4.8 \\
Trial 3         & 2.3     & 2.1     & 3.8     & 3.3     & 4.3 \\
Trial 4         & 2.5     & 2.0     & 3.0     & 2.5     & 4.0 \\
Trial 5         & 2.0     & 2.0     & 3.5     & 4.0     & 5.0 \\
Trial 6         & 2.0     & 2.0     & 3.8     & 4.0     & 5.5 \\
Avg. Error      & 1.21    & 0.25    & 0.48    & 0.53    & 0.48\\ \bottomrule
\end{tabular}
\end{table}

\subsection{Running Stitch Experiment}
\begin{figure}[tb]
	\centering
	\includegraphics[width=\linewidth]{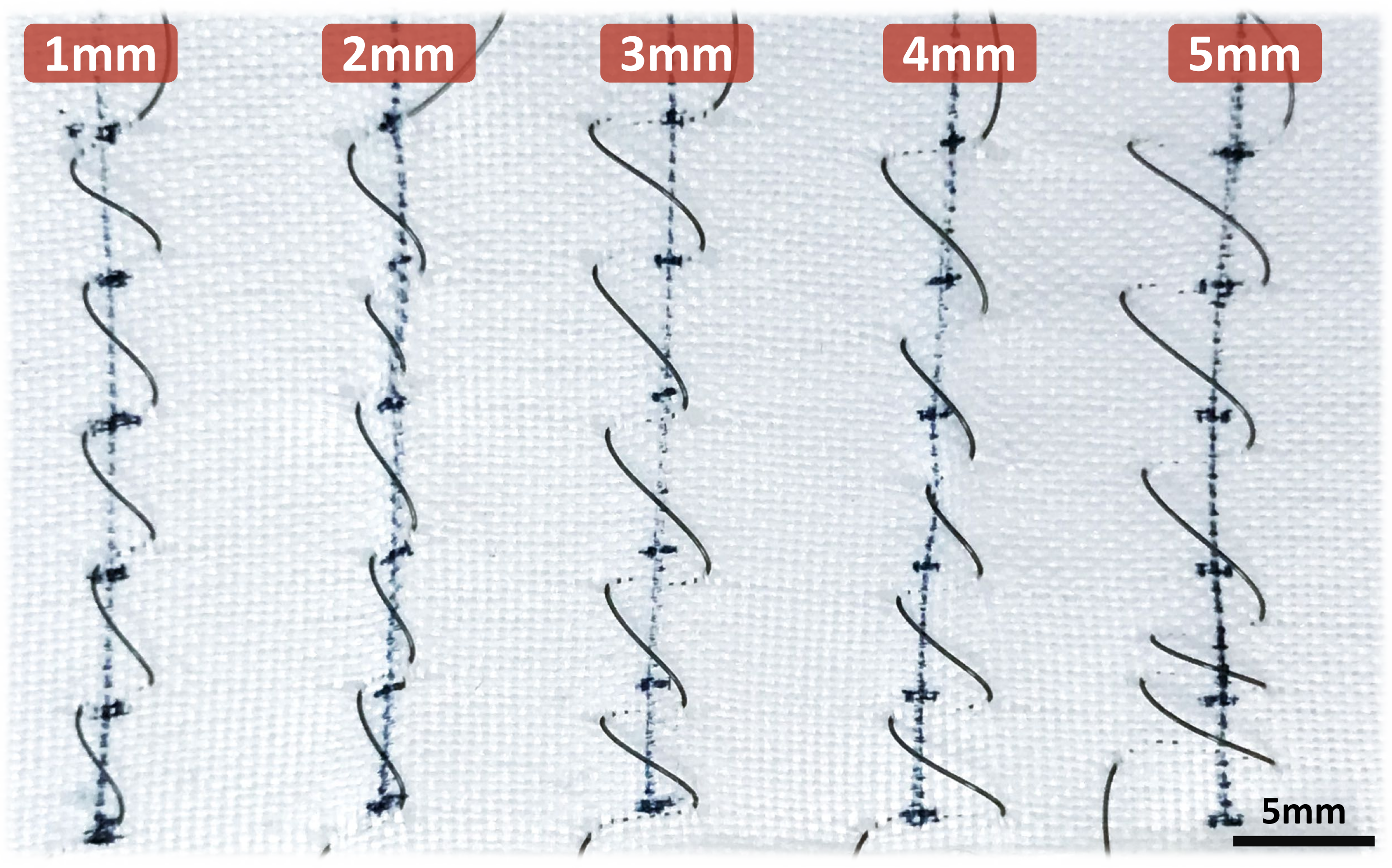}
	\caption{The results of the running stitch task performed by the robot autonomously. Each column represents a trial with various desired stitch size ranging from 1mm to 5mm.}
	\label{fig:stitch_exp}
\end{figure}

To quantitatively validate the accuracy and robustness of the stitching process, a running stitch experiment is designed. As shown in Fig.~\ref{fig:stitch_exp}, a line is drawn on a sheet of fabric to indicate the desired stitch targets. To test the optimal stitch size the system is capable of, the experiment consists of 5 trials where each trial uses different stitch size $d$ ranging from 1mm to 5mm. At the beginning of each trial, the user selects six stitching targets along the line. For each stitch, a trajectory is planned based on the local 3D profile and the stitch size $d$. The robot follows the planned trajectory by using the visual servoing technique. After each stitch is accomplished, the robot moves to a predefined standby pose before carrying the next stitch. 

We measure the distance between the entry and exit points of the thread on the fabric as the actual stitch size. The quantitative and qualitative results of the experiment are shown in Table \ref{tb:stitch} and Fig.~\ref{fig:stitch_exp}, respectively. The error for each trial is calculated as the mean absolute difference between the desired and actual stitch size.
Most trials achieve an error less than 0.6mm except the 1mm trial which has an average error of 1.2mm. The actual stitch size for the 1mm group is always not less than 2mm which leads to a consistent error. The main reason of the error is that as the sewing device closes the jaws to pass the needle in the fabric (as shown in step 4 of Fig.~\ref{fig:stitch_steps}), the left side of the jaw would press and deform the fabric when the needle comes out of the fabric.

\subsection{Knot Tying for Stent Graft}

The performance of the knot tying method is evaluated by sewing one stent on the graft. In order to perform sewing on the surface of the stent graft, a cylindrical object, called a mandrel, is used for holding the graft and the stent in place during sewing. This mandrel provides tension for the fabric being sewn. The area under the stent graft is hollow so that the needle can easily pierce through without contacting with the mandrel. The thickness, edge length and the diameter of the stent used in this experiment is 0.4mm, 12mm and 22mm respectively.

The knots on the stent are completed successively where each knot consists of a stitch with an overhand knot. Similar to the running stitch experiment, the user selects the desired knot position on the sewing target, which is along the stent edge in this case. The tail of the thread is attached to a fixed position before starting using our method to successively apply stitch and knot. Once the robot finishes the stitch, it moves the sewing device to the standby pose followed by the knot tying step.
In this paper, we used a nylon thread with a diameter of 0.2mm which has been used for commercial handmade stent graft. The desired force to keep the thread tension is 0.7N along the dominant axis while the force required to secure a knot is around 2N. The length of the thread used for the experiment is approximated 250mm, which could be applied for tying 12-16 knots using our method. 
The execution time of the vision-guided stitching and tying a single knot is about 3 minutes and 2 minutes, respectively. See the supplementary video via \url{https://youtu.be/WX63S55PE3w}.

\begin{figure}[tb]
	\centering
	\includegraphics[width=\linewidth]{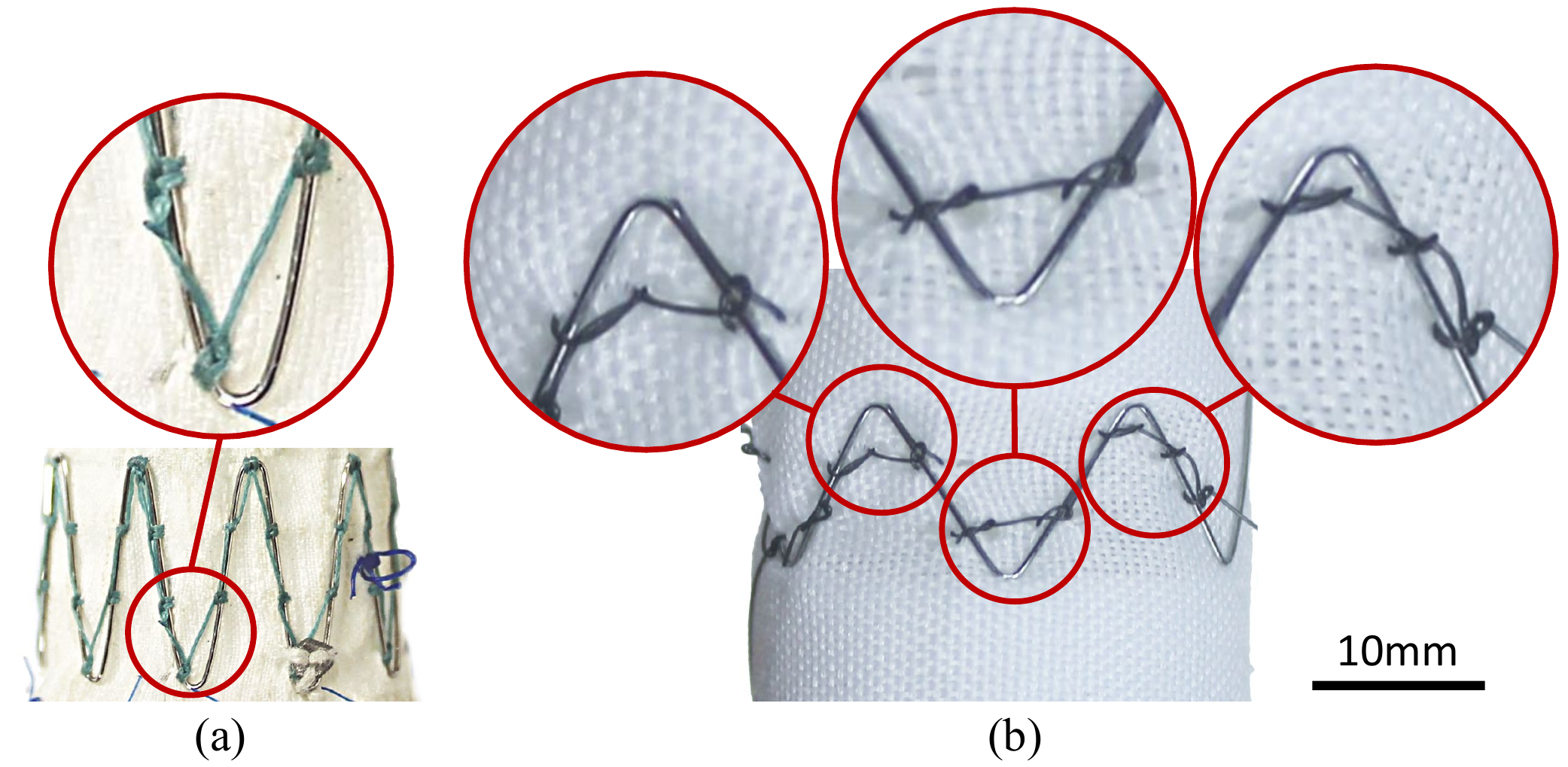}
	\caption{(a) An example of commercial handmade stent graft. (b) The results of the sewed stent produced by the proposed autonomous knot tying method.}
	\label{fig:stent_result}
\end{figure}

\section{Discussion}
\label{sec:discuss}

In this paper, we have proposed a novel single sided 3D sewing system for arbitrary 3D structures. For the stitch size planning method, sub-millimetre accuracy is achieved. Due to the geometry constraints of the circular needle, the sewing device is not suitable for making a stitch whose size is larger than 5mm. It is worth noting that making a small stitch less than 2mm needs to consider deformation of the fabric. To solve this problem, a new mechanism which could precisely reorientate the sewing device is helpful. Then the robot arm only serves for positioning purpose. In addition, we found the tension on the fabric is another factor which can influence the sewing accuracy. The needle may fail to enter the fabric or stitch a wrong position if the fabric is not well tensioned.

The accuracy of the needle positioning using the visual servoing technique depends on several factors. The point cloud of the sewing targets obtained from the depth camera can be noisy as the fabric surface is slanted in the camera view. To address this issue, the fabric can be brought towards the camera which also requires modification of the pattern attached to the device as it needs to be visible for the visual servoing. Another limitation of the vision system is the resolution of the depth camera which is only 640x480. When the user selects the target on the image, the 3D position of adjacent pixels can be very far. Nevertheless, this can be addressed as the hardware is upgraded (the latest RealSense\textsuperscript{TM} D-series depth camera supports 1280x720 resolution). Furthermore, a useful improvement for the proposed system is the markerless tracking of the device or needle. This will extend the flexibility of the system by allowing control of the sewing device from different viewing angles and occlusion.

For the knot tying experiment, we have demonstrated a promising solution for applications requires successively applying knot to secure each stitch. This could secure the stent more reliably compared with purely using running stitch. During the experiment, some issues were also identified. First, to keep the tension on a long thread, the arm holding the thread manipulator needs to move a long distance, which is not cost-effective and may reach its workspace limitation. A rolling mechanism which could automatically wind the thread can be a potential solution. Second, the overhand knot made by the robot is tied around an object. The knot has the working end brought over and under the standing end. It can also be called a half hitch. To better secure each stitch, it is desired to apply two successive half hitches to make up the common clove hitch. Third, the direction and force to pull the thread to secure the knot are important to make a proper knot. As shown in Fig.~\ref{fig:stent_result}, the robot made knot is not as tight as the handmade knot. The thread used in our experiment is more elastic which can be loose even it is pulled tightly. However, another more important reason is that it is not the optimal directions for pulling the thread to place the knot. Kang et al. \cite{kang2002robotic} has proposed a knot sliding condition for properly placing a knot, which can be further investigated in future.

% (1) Visual Servoing Accuracy and How to improve, Local camera, model-based servoing, servoing needle?
% (2)sewing point detection, stent recognition, could from CAD, or from patient geometry 
% (3) Various Stitch sizes Accuracy and how to improve, mechanically? 
% (4) Automatic knot tying. Thread management. A roller to wind thread, so the thread would not fly everywhere

\section{Conclusions}

In this paper, we have proposed an autonomous robotic sewing system with an integration of the ABB YuMi\textsuperscript{\textregistered} dual-arm robot, closed-loop visual servoing control and the customised sewing hardware for personalized stent graft manufacturing. The customised sewing device facilitates the automation of the sewing task by providing needle handling on one hand.
We have used closed-loop visual servoing based on RGBD sensing to achieve accurate device positioning and steady motion which benefits the autonomous stitching. Leveraging on this novel system, a trajectory planning method for the sewing various stitch sizes is proposed. In addition, a novel thread manipulator with force sensing capability is proposed for thread manipulation during knot tying. The thread manipulator allows simplified thread manipulation during the robotic sewing. We have demonstrated that the proposed system can accomplish continuous stitches with reasonable accuracy as well as successively tie overhand knots for the stent graft manufacturing. The system has shown the potential to be transferred to other clinical areas such as surgical suturing.

% \addtolength{\textheight}{-12cm}   
% This command serves to balance the column lengths
% on the last page of the document manually. It shortens
% the textheight of the last page by a suitable amount.
% This command does not take effect until the next page
% so it should come on the page before the last. Make
% sure that you do not shorten the textheight too much.

%%%%%%%%%%%%%%%%%%%%%%%%%%%%%%%%%%%%%%%%%%%%%%%%%%%%%%%%%%%%%%%%%%%%%%%%%%%%%%%%

%%%%%%%%%%%%%%%%%%%%%%%%%%%%%%%%%%%%%%%%%%%%%%%%%%%%%%%%%%%%%%%%%%%%%%%%%%%%%%%%

%%%%%%%%%%%%%%%%%%%%%%%%%%%%%%%%%%%%%%%%%%%%%%%%%%%%%%%%%%%%%%%%%%%%%%%%%%%%%%%%

\section*{ACKNOWLEDGMENT}
This work is supported by UK EPSRC Personalised Stent
Graft Manufacturing for Endovascular Intervention Project
(EP/L020688/1). The authors would like to thank Yoshua Nava from KTH for his assistance on the YuMi robot.

\bibliographystyle{myIEEEtran}
\bibliography{mybib}

\end{document}